\documentclass[runningheads]{llncs}
\usepackage[T1]{fontenc}
\usepackage{amssymb}
\usepackage{booktabs}
\usepackage{comment}
\usepackage{graphicx}
\usepackage{hyperref}
\usepackage{multirow}

\newcommand{\etal}{\textit{et al.}}

\begin{document}

\title{ELMTEX: Fine-Tuning Large Language Models for Structured Clinical Information Extraction. A Case Study on Clinical Reports}

\titlerunning{Fine-Tuning LLMs for Structured Clinical Information Extraction}

\author{Aynur Guluzade\inst{1}\orcidID{0000-0002-4410-1095} \and
Naguib Heiba\inst{1}\orcidID{0000-0002-3657-970X} \and
Zeyd Boukhers\inst{1}\orcidID{0000-0001-9778-9164} \and
Florim Hamiti\inst{1}\orcidID{0000-0001-5235-4423} \and
Jahid Hasan Polash\inst{1}\orcidID{0009-0000-0099-9238} \and
Yehya Mohamad\inst{1}\orcidID{0000-0002-6240-0099} \and
Carlos A Velasco\inst{1}\orcidID{0000-0001-8342-4471}
}

\authorrunning{Guluzade \etal}

\institute{Fraunhofer Institute for Applied Information Technology FIT\\
Schloss Birlinghoven 1, 53757 Sankt Augustin, Germany\\
\email{\{aynur.guluzade,naguib.heiba,zeyd.boukhers,florim.hamiti,\\
jahid.hasan.polash,yehya.mohamad,\\
carlos.velasco.nunez\}@fit.fraunhofer.de}\\
\url{https://www.fit.fraunhofer.de/}}

\maketitle

\begin{abstract}
Europe's healthcare systems require enhanced interoperability and digitalization, driving a demand for innovative solutions to process legacy clinical data. This paper presents the results of our project, which aims to leverage Large Language Models (LLMs) to extract structured information from unstructured clinical reports, focusing on patient history, diagnoses, treatments, and other predefined categories. We developed a workflow with a user interface and evaluated LLMs of varying sizes through prompting strategies and fine-tuning. Our results show that fine-tuned smaller models match or surpass larger counterparts in performance, offering efficiency for resource-limited settings. A new dataset of $60,000$ annotated English clinical summaries and $24,000$ German translations was validated with automated and manual checks. The evaluations used ROUGE, BERTScore, and entity-level metrics. The work highlights the approach's viability and outlines future improvements.


\end{abstract}

\section{Introduction}
There is a growing need for interoperability and digitalization in Europe \cite{EuropeanCommission.2019}. The variety of health systems and the need to cope with legacy documentation and procedures present in the sector foster the search for innovative approaches, which should address the upcoming requirements of the European Health Data Space (EHDS \cite{ehds.ep.2022,ehds-etri:2022}). Recent advances in Large Language Models (LLMs) offer a potential for structured clinical information extraction (IE) to enhance the quality and interoperability of healthcare. Furthermore, automated IE reduces the manual effort required by healthcare professionals in daily tasks, such as clinical decision-making, research, and operational procedures, which require a rapid extraction of structured and standardized information, addressing the specialized language and terminologies used in medical documents \cite{landolsi2023information}.

The work presented in this paper focuses on the optimization of LLMs to extract structured and standardized information from text-based clinical reports. This work is part of an internal wider project of our Institute in collaboration with clinical teams called ELMTEX, which included a workflow allowing the processing of legacy documentation, sometimes available as images of text documents, which were scanned and automatically transformed into text information, and the development of a user interface that allows medical teams to validate the extracted data. These additional components are not addressed in this paper.

We optimized the models to extract information on multiple categories. We evaluated open-source LLMs of various sizes, first without fine-tuning by only prompting them, and subsequently fine-tuning smaller ones (Llama 3.1 8B, Llama 3.2 1B \& 3B \cite{dubey2024llama}) to observe their performance improvement. We assessed the impact of fine-tuning on the accuracy of IE from clinical reports using different metrics, which include a wide range of evaluation metrics, from n-gram-based to similarity-based and entity-level metrics. We explored methods ranging from basic prompting to advanced prompting with in-context learning, as well as LoRA fine-tuning. Experiments indicated that fine-tuning Small Language Models (SLMs) achieves better results compared to existing large models.

To support these experiments, we introduced a new large corpus comprising $60,000$ annotated clinical reports in English and $24,000$ in German. Each instance includes a summary extracted from PubMed Central articles\footnote{\url{https://pmc.ncbi.nlm.nih.gov/}} and its corresponding structured information in JSON format. The dataset spans diverse categories, including patient family/social history, medical history, diagnosis, and outcome assessment, ensuring comprehensive coverage of key medical information. To facilitate reproducibility and reuse, our code implementation\footnote{\url{https://gitlab.cc-asp.fraunhofer.de/health-open/elmtex}} and the dataset \cite{Guluzade.Dataset.2025} are publicly available.

\section{State-of-the-Art}
Earlier approaches for data extraction from clinical reports include rule-based systems~\cite{mykowiecka2009rule}. New approaches involve supervised machine learning (ML) models \cite{adamson2023approach,hahn2020medical}. The constant evolution in the medical field requires the update of such methods, implying a manual effort \cite{adnan2019limitations}. Likewise, supervised ML models also need large annotated datasets, which are expensive and take a long time to create \cite{zajkac2023ground}.

The rising development of Large Language Models (LLMs) in recent years has shown significant potential in Natural Language Processing (NLP), including clinical IE. Transformer-based LLMs \cite{vaswani2017attention} can handle and understand large amounts of text with limited task-specific training without the need for a large labeled dataset \cite{radford2019language}. A concern is that these models are pre-trained on general texts and they often lack domain-specific knowledge. This can cause LLMs to hallucinate and generate plausible but inaccurate information because the model faces difficulties in understanding medical reports terminology \cite{ji2023survey}.

Clinical text has a different syntax and vocabulary compared to general text \cite{wu2020deep}, leading to the development of domain-specific models. Early efforts included clinical word embeddings \cite{wu2015clinical}, and later models such as ClinicalBERT, SciBERT, BioBERT, and PubMedBERT were inspired by BERT \cite{devlin2018bert}. However, some studies report only minimal performance gains over classical methods such as random forest \cite{chen2020intimate}, and LLMs like GPT-3 struggle to achieve competitive results in biomedical NLP tasks \cite{moradi2021gpt}. Recent studies have shown that LLMs such as GPT-3.5 and GPT-4 exhibit strong performance in zero- and few-shot learning scenarios for clinical IE tasks, even without domain-specific training \cite{hu2024improving}. 

However, concerns persist about their accuracy and potential to generate plausible yet incorrect information (hallucinations \cite{pal2023med}). To address these challenges, researchers have explored fine-tuning LLMs on domain-specific data. For example, Gema~\etal \cite{gema2023parameter} introduces a two-step parameter-efficient fine-tuning framework for LLMs in clinical applications; Wang~\etal \cite{wang2023clinicalgpt} presents a language model optimized for clinical scenarios through fine-tuning with real-world medical data; and Peng~\etal \cite{peng2024model} evaluate soft prompt-based learning algorithms for LLMs in clinical concepts and relation extraction.

All these approaches rely on the availability of high-quality datasets which is crucial for advancing the task. Open access electronic health record (EHR) datasets, such as those provided by the MIMIC database, have been instrumental in developing and evaluating NLP models \cite{nuthakki2019natural}. However, the scarcity of large-scale annotated clinical datasets, particularly for non-English languages, remains a significant barrier. Efforts are underway to create synthetic datasets and translate existing ones to mitigate this \cite{roller2022medical}.

\section{ELMTEX Dataset and Evaluation Approach}
For a clinical report $R$, represented as a sequence of tokens $R = \{r_1, r_2, \dots, r_n\}$, the task is to map $R$ to a structured representation $S$: 
\begin{equation}
    S = \{C_1: s_1, C_2: s_2, \dots, C_k: s_k\}
\end{equation}

\noindent where $S$ is an object containing predefined categories; $C = \{C_1, C_2, \dots, C_k\}$ represents the set of categories with $k = 15$; and $s_i$ is the extracted information for category $C_i$, expressed as a string. For each $s_i$, we use separate concepts delimited by semicolons ($;$) since each category may contain multiple concepts. The categories $C_i$ are listed in Table~\ref{tab:stats}. 

The goal is to train and evaluate LLM $f_\theta$, to approximate the mapping function $f_\theta : R \to S$, such that for each category $C_i$, the extracted string $s_i$ accurately reflects the corresponding structured information derived from $R$. The dataset $\mathcal{D}$ used for training and evaluation consists of $N$ samples, where each sample $(R, S^\ast)$ includes a clinical report $R$ and its corresponding ground truth structured representation $S^\ast$.

\subsection{Evaluation}
The experiments explore three model setups, described in the following:

\subsubsection{Naive Prompting.}
It consists of directly querying the LLM with a simple instruction to extract information for all categories from $R$. We constructed a simple prompt $P$ to describe the task and explicitly list the categories $C$ from which information needs to be extracted, without providing detailed definitions or scope for each category. The expectation is that the LLM $f_\theta$ can infer the meaning and scope of each category based solely on the category names and perform the task as instructed.

This setup relies on the LLM's inherent ability to understand the semantic meaning of the task described in the prompt, comprehend the intent behind each category name, and generate syntactically valid and semantically accurate structured output without further guidance. However, due to the lack of explicit definitions or examples for each category, the model's output may vary significantly in quality and completeness. The performance in this setup depends primarily on the model's pre-training data and generalization capabilities.

\subsubsection{Advanced Prompting with In-Context Learning.}
It includes examples of input-output pairs as context within the prompt to guide the LLM. Here, we explicitly defined each category $C_i \in C$, providing clear descriptions and the scope of information expected for each category. To further improve task performance, we integrated in-context learning by incorporating examples retrieved from a training set. Given a clinical report $R$ from the test set, an encoder-based retrieval model $g_\phi$ retrieves the top $m$ most similar clinical reports $\mathcal{R}' = \{R_1, R_2, \dots, R_m\}$ from the training set, where the training set is disjoint from the test set. These retrieved reports are paired with their corresponding annotated structured representations $\mathcal{S}' = \{S_1^\ast, S_2^\ast, \dots, S_m^\ast\}$, forming the in-context examples. Formally, the retrieval process is defined as:

\begin{equation}
    g_\phi : R \to \{(R_1, S_1^\ast), (R_2, S_2^\ast), \dots, (R_m, S_m^\ast)\}
\end{equation}

\noindent where $g_\phi$ ranks training examples based on their semantic similarity to $R$ using cosine similarity. The prompt $P$ is constructed with a detailed task description, instructing the LLM to extract structured information from $R$. The explicit definitions and scopes for all categories $C$ ensure clarity in the expected output for each category. In addition, the retrieved clinical reports $\mathcal{R}'$ and their corresponding annotated structured representations $\mathcal{S}'$ are appended to the prompt as in-context examples, demonstrating how the task should be performed. Our objective is to leverage the LLM to use explicit task instructions and category definitions to better understand the task. We intended that the LLM learns from the retrieved in-context examples to handle the clinical report of the test $R$ more effectively and to generalize to new instances using similar previous examples. By integrating retrieval-augmented in-context learning, this setup mitigates ambiguities in naive prompting and enables the LLM to produce a more accurate and reliable output for each category.

\subsubsection{LLM Fine-tuning.}
We trained small LLMs on a domain-specific dataset to optimize $f_\theta$ for clinical reports. Building upon our naive prompting approach, we employed parameter-efficient fine-tuning using Low-Rank Adaptation (LoRA) \cite{hu2021lora}. This method enables the LLM $f_\theta$, to adapt to the task while maintaining its general pre-training capabilities. The goal is for the LLM to learn the task-specific mapping $f_\theta: R \to S$ and the definitions and scope of each category $C_i \in C$ during fine-tuning, therefore eliminating the need for detailed prompts at inference time. Fine-tuning is conducted on the training set, which comprises 90\% of the total dataset $\mathcal{D}$. Each training instance consists of a clinical report $R$ and its corresponding ground truth structured representation $S^\ast$. The objective of fine-tuning is to minimize the loss $\mathcal{L}$ between the predicted structured representation $S$ and the ground truth $S^\ast$:

\begin{equation}
    \mathcal{L} = \frac{1}{N} \sum_{i=1}^N \textrm{Loss}(f_\theta(R_i), S_i^\ast)
\end{equation}

\noindent where $N$ is the number of training samples, and \textrm{Loss} refers to the loss of cross-entropy. The fine-tuning process enables the LLM to internalize the task-specific knowledge required, learn the definitions and scopes of each category $C_i$ directly from the training data, and generalize effectively to unseen test samples, including edge cases and ambiguous scenarios. 

By fine-tuning on domain-specific data, the model becomes more robust and precise, overcoming the limitations of relying solely on prompt engineering. This is particularly important given the complexity and variability of clinical reports, where complete coverage of all possible scenarios through prompts alone is unfeasible. Unlike the other setups, which depend heavily on prompt construction, the fine-tuned LLM requires only minimal instructions at inference time.

\subsection{Dataset Generation Workflow}
We introduced a new dataset of clinical report summaries, annotated with structured information across $15$ categories \cite{Guluzade.Dataset.2025}. This dataset was created to address the lack of large-scale resources for clinical IE. It also promotes the development of methods tailored to clinical data, helping to improve healthcare provision. The dataset contains $60,000$ annotated English clinical report summaries, from which we translated over $24,000$ examples into German. The dataset is based on PMC-Patients \cite{Zhao.2023b}, a collection of $167,000$ patient summaries from case reports in PubMed Central. We extracted a subset of patient reports for our work and used a semi-automated approach to create the dataset. First, we defined the categories and their scopes by reviewing related work \cite{caufield2018reference} and consulting physicians to ensure that the selected categories were relevant. Afterwards, we manually annotated reports to serve as in-context learning examples. We then used the GPT-4 model with advanced prompting and in-context learning to generate the initial annotations.

Table~\ref{tab:stats} provides the dataset statistics for each category, including error rates, derived from samples used for manual validation. Note that not all 15 categories are present in every patient report; their presence depends on the details of the report. To reflect this, we also include the percentage of reports in which each category is present. 
Please refer to our Appendix \cite{Guluzade.Appendix.2025} for examples of the prompts used, and the results on the German dataset.

\begin{table}[t]
\centering
\caption{Annotated categories with their presence percentages and corresponding error rates in the English dataset.}
\label{tab:stats}
\begin{tabular}{l|c|c}
    \toprule
    \textbf{Category} & \textbf{Presence (\%)} & \textbf{Error Rate (\%)} \\
    \midrule
    $ C_1 $: \textit{age} & 100 & 0.00 \\
    $ C_2 $: \textit{comorbidities} & 37.47 & 11.00 \\
    $ C_3 $: \textit{diagnosis} & 98.63 & 8.00 \\
    $ C_4 $: \textit{diagnostic\_procedures} & 98.87 & 1.67 \\
    $ C_5 $: \textit{family\_history} & 17.86 & 0.67 \\
    $ C_6 $: \textit{gender} & 100 & 0.00 \\
    $ C_7 $: \textit{interventional\_therapy} & 73.30 & 4.17 \\
    $ C_8 $: \textit{laboratory\_values} & 67.75 & 2.67 \\
    $ C_9 $: \textit{life\_style} & 22.70 & 7.00 \\
    $ C_{10} $: \textit{medical\_surgical\_history} & 84.29 & 7.50 \\
    $ C_{11} $: \textit{pathology} & 73.56 & 4.67 \\
    $ C_{12} $: \textit{patient\_outcome\_assessment} & 92.88 & 1.00 \\
    $ C_{13} $: \textit{pharmacological\_therapy} & 70.33 & 1.50 \\
    $ C_{14} $: \textit{signs\_symptoms} & 95.96 & 2.33 \\
    $ C_{15} $: \textit{social\_history} & 7.94 & 7.00 \\
    \bottomrule
\end{tabular}
\end{table}

\section{Experiments}
\subsection{Experimental setup}

\subsubsection{Baseline Models}
We used the Llama 3 series \cite{dubey2024llama} of models for our experiments. For small-sized models, we employed Llama 3.2 1B\&3B Instruct for advanced prompting with in-context learning and LoRA fine-tuning setups. We skipped naive prompting for these models, as their size and pre-training were insufficient for effectively performing the task or generating properly formatted JSON outputs. For medium-sized models, we used Llama 3.1 8B Instruct across all setups, including naive prompting, advanced prompting, and fine-tuning. For large models, we selected Llama 3.1 70B Instruct and Llama 3.1 405B Instruct. These were tested only with naive and advanced prompting, as fine-tuning such large models is not optimal for a task of this specificity. For the Llama 405B Instruct model, we used FP8 quantization due to GPU resource limitations. The experiments were primarily run on 4 H100 GPUs. However, smaller and medium-sized models could be run on a single GPU, as LoRA fine-tuning reduces the GPU memory requirements.

\subsubsection{Evaluations Metrics}
We used three metrics for evaluation, chosen to provide complementary insights into the accuracy and relevance of the extracted information. We evaluated each category independently, averaging the results across all categories at the end.
First, we used ROUGE to measure the n-gram overlap (unigrams, bigrams, and longest common subsequence) between the model's output and reference summaries, assessing content and structural alignment.
Second, BERTScore was used to evaluate the semantic similarity using contextual embeddings. We calculated precision, recall, and F1 scores based on cosine similarity, to ensure that the generated summaries captured the intended meaning.
Finally, we considered the Entity-level metrics, focused on clinical accuracy by extracting entities such as medications and diagnoses with a medical NER model\footnote{\url{https://github.com/allenai/scispacy}}. We calculated precision, recall, and F1 scores to compare the extracted entities between the outputs and references.

Each of these metrics was chosen to target specific aspects of the evaluation: ROUGE for surface-level alignment, BERTScore for semantic understanding, and entity-level metrics for domain-specific accuracy. This combination provided a well-rounded assessment of the performance of the model.

\subsection{Results}
Table~\ref{tab:results} summarizes the results comparing different setups in the models. We observe that the LLama 3.1 8B fine-tuned model achieves the best overall performance across all metrics, outperforming all non-fine-tuned models, including the LLama 405B, even with advanced prompting. This suggests that fine-tuning is crucial for enabling the model to fully grasp the various definitions and concepts required for each category. We also see that fine-tuning smaller models like LLama 3.2 1B\&3B yields surprisingly strong results. These models, which can run on edge devices and require far fewer resources, demonstrate their potential for practical deployment in resource-constrained environments.

On the other hand, we note that only large models perform well with naive prompting, which confirms their inherent advantage due to their scale and pre-training. Advanced prompting and in-context learning, however, enable medium and large models to perform significantly better, underscoring the value of this more sophisticated approach.

\begin{table}
\centering
\def\arraystretch{1.1}
\caption{Model performance comparison on our English dataset.}
\label{tab:results}
\begin{tabular}{cl|ccc|c|ccc}
\toprule
& \multirow{2}{*}{\textbf{Models}} & \multicolumn{3}{c|}{\textbf{ROUGE}} & \textbf{BERTSc.} & \multicolumn{3}{c}{\textbf{Entity-Level}} \\
& & R-1 & R-2 & R-L & F1 & P & R & F1 \\ \midrule
\multirow{3}{*}{\rotatebox{90}{Naive}} 
& Llama-3.1-8B-Instr. & 0.5585 & 0.4303 & 0.5432 & 0.5949 & 0.6592 & 0.5548 & 0.6025 \\
& Llama-3.1-70B-Instr. & 0.5837 & 0.4772 & 0.5989 & 0.6359 & 0.6773 & 0.5833 & 0.6267 \\
& Llama-3.1-405B-Instr. & 0.6183 & 0.5137 & 0.6261 & 0.6696 & 0.7012 & 0.6025 & 0.6481 \\ \midrule
\multirow{5}{*}{\rotatebox{90}{Adv.+ICL}} 
& Llama-3.2-1B-Instr. & 0.2333 & 0.1558 & 0.2255 & 0.3396 & 0.4298 & 0.2564 & 0.3211 \\
& Llama-3.2-3B-Instr. & 0.4112 & 0.3105 & 0.3917 & 0.5553 & 0.5879 & 0.4861 & 0.5321 \\
& Llama-3.1-8B-Instr. & 0.6244 & 0.5178 & 0.6033 & 0.6986 & 0.6818 & 0.6872 & 0.6844 \\
& Llama-3.1-70B-Instr. & 0.6993 & 0.5865 & 0.6792 & 0.7379 & 0.7099 & 0.7392 & 0.7242 \\ 
& Llama-3.1-405B-Instr. & 0.6969 & 0.5716 & 0.6714 & 0.7287 & 0.7312 & 0.7407 & 0.7359 \\ \midrule
\multirow{3}{*}{\rotatebox{90}{Fine-t.}} & Llama-3.2-1B-Instr. & 0.7416 & 0.6350 & 0.7247 & 0.7857 & 0.7595 & 0.7502 & 0.7548 \\
& Llama-3.2-3B-Instr. & \underline{0.7721} & \underline{0.6727} & \underline{0.7555} & \underline{0.8122} & \underline{0.7853} & \textbf{0.7840} & \underline{0.7846} \\
& \textbf{Llama-3.1-8B-Instr.} & \textbf{0.7771}  & \textbf{0.6841}  & \textbf{0.7626} & \textbf{0.8253} & \textbf{0.7917} & \underline{0.7822} & \textbf{0.7869} \\
\bottomrule
\end{tabular}
\end{table}

\subsection{Error Analysis}
For the error analysis, we randomly sampled 20 incorrect predictions for each model experiment setup and examined the types of errors. The analysis revealed two main error types:

\begin{itemize}
    \item \textbf{Missing Extracted Information} Models often missed specific concepts within certain categories. This was particularly common in naive and advanced prompting setups. While the main concepts for each category were usually extracted, additional relevant concepts were often missed. This error was mainly observed in categories like diagnostic\_techniques\_procedures, diagnosis, laboratory\_values, and pharmacological\_therapy. A possible reason is that these categories can involve lengthy lists of concepts, which can be spread over multiple sentences in clinical reports, making it challenging for the LLM to capture all relevant details.
    \item \textbf{Wrongly Categorized Concepts} Errors involving the misplacement or incorrect categorization of concepts were observed, particularly between similar categories. This issue was more frequent in naive and advanced prompting setups and was the primary error type for fine-tuned smaller LLMs (1B\&3B). The affected categories included life\_style, family\_history, social\_history, medical\_surgical\_history, signs\_symptoms, and comorbidities. Such errors can occur due to misinterpretation of the LLMs in sentences.\newline
    Additionally, for LLama 3.2 1B with advanced prompting, we identified instances of complete hallucination, where the model generated non-existent concepts or concepts copied from in-context learning examples. This behavior was expected given the small size of the model and its limited capacity to retain pre-trained knowledge and follow detailed instructions.
\end{itemize}

\section{Conclusions and future work}
In this article, we investigated the potential of LLMs for clinical IE, focusing on tasks involving structured data generation from unstructured clinical reports. We systematically evaluated approaches like naive prompting, advanced prompting with in-context learning, and fine-tuning across LLMs of varying sizes, and observed that fine-tuning not only enhances performance but also bridges the gap between large and small LLMs. The results highlight the practical effectiveness of smaller fine-tuned models for deployment in real-world clinical settings with limited resources. To support our experiments, we released a large-scale clinical dataset containing $60,000$ patient summary reports in English. The dataset offers extensive coverage of various clinical categories and serves as a valuable benchmark in the field.

Our future work is focused on different aspects. First, we are refining our demonstrator user interface to facilitate clinicians the evaluation of the results of the model. We also identified that the categories need to be refined to be better synchronized with the standard terminologies and data models in the health sector, as highlighted by some of the approached clinical teams. This will contribute to the refinement of the domain-specific training. Additionally, we are considering the expansion of the multilingual capabilities of the dataset and the training approach. Furthermore, we should also consider how our developments are influenced by the AI-Act \cite{AI-Act:2024} in Europe.


\end{document}